\documentclass[%
 aip,
 amsmath,amssymb,
 reprint,%
]{revtex4-1}

\usepackage{siunitx}
\usepackage{graphicx}
\usepackage{algorithm}
\usepackage{algpseudocode}
\usepackage{subcaption}
\usepackage{svg}
\usepackage{graphicx}
\usepackage{dcolumn}
\usepackage{bm}

\begin{document}

\preprint{AIP/123-QED}

\title{Learning swimming via deep reinforcement learning
}

\author{Jin Zhang}
\author{Lei Zhou}
\author{Bochao Cao}%
 \email{Corresponding author: cbc.fudan.edu.cn}
 \affiliation{Department of Aeronautics and Astronautics, Fudan University, Shanghai 200433, China}

\date{\today}

\begin{abstract}
In this paper, reinforcement learning (RL) method is used to optimize the efficiency of a flapping foil in a water tunnel environment. In the optimization process, the foil continuously adjusts its motion based on the feedback collected from the environment. Several motion patterns are obtained from the experiment which can result in high hydrodynamic performance compared to pure harmonic motions. The length of the efficiency evaluation window is found to be crucial when optimizing the long-term efficiency of the flapping foil. Our results demonstrate the feasibility of using RL method for optimizing hydrodynamic performance in a real-world fluid environment.
\end{abstract}

\maketitle

\section{\label{sec:level1}INTRODUCTION}

The study of oscillating foils is a classic field in fluid mechanics, inspired by the locomotion paradigm of fish and aquatic mammals. By generating large-scale vortices through body motion and manipulating them to enhance hydrodynamic performance, oscillating foils hold great promise for the design of new aquatic propulsion technologies and for understanding animal locomotion.

Extensive literature vividly describes the behavior of foils under sinusoidal motion\cite{TriantafyllouM.S2000HoFS,WuTheodoreYaotsu2011Fsab,SmitsAlexanderJ.2019Uaos}. Studies on the effects of the non-dimensional frequency, the non-dimensional amplitude of motion, and the Reynolds number on hydrodynamic performance and wake structure have been conducted both numerically and experimentally \cite{SenturkUtku2018Nsot,SenturkUtku2019RNSo,ANDERSONJ.M.1998Ofoh,MackowskiA.W.2015Dmot}, with scaling laws proposed to predict propulsive performance or wake transition under sinusoidal motion\cite{FloryanDaniel2017Stpp,LagopoulosN.S.2019Uslf}.

Moreover, some studies have investigated the influence of intermittent actuation and non-sinusoidal gaits, suggesting that changing the sinusoidal motion pattern may yield potential efficiency benefits. Intermittent swimming is always energetically favorable, but usually accompanied by a loss of speed, and an optimal duty cycle for energy savings exists\cite{FloryanDaniel2017Faeo,AkozEmre2018Upba}. Non-sinusoidal gaits also have an obvious influence on propulsive performance. Square-like gaits exhibit much higher thrust and power than triangular-like and sinusoidal ones, while the highest efficiency is always achieved by sinusoidal gaits\cite{WOS:A1989AM56000008,LuK.2013Nsol,VanBurenT.2017Ngfu,MuhammadZaka2022Eteb}. Compared with continuous sinusoidal motion, non-sinusoidal flapping motion includes a much wider range of motion patterns, which can generate more complex flow structures that might benefit the propulsion efficiency or other aspects of a swimming body. This makes non-sinusoidal motion a better candidate in controlling flow fields around a flapping foil.

\begin{figure*}
  \centering
\includegraphics[width=0.9\textwidth]{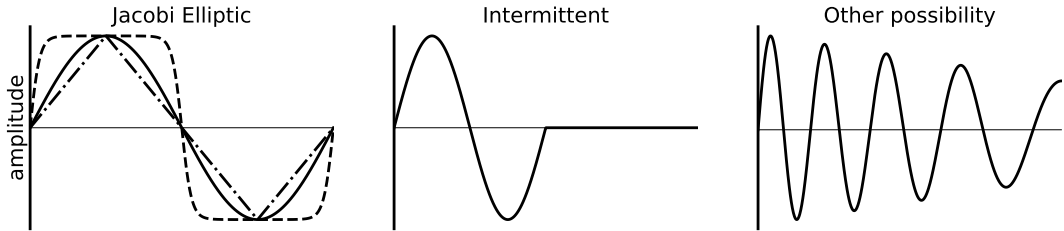}
  \caption{Illustration of non-sinusoidal motions.}
  \label{fig:waveforms}
\end{figure*}
In previous studies on non-sinusoidal motions, motion patterns are typically prescribed as certain periodic functions to simplify implementation and analysis. Consequently, locomotion tends to be constrained within certain subdomains of the whole motion space. This could potentially exclude other possible efficient motion patterns in the investigation, as illustrated in Fig. \ref{fig:waveforms}. In this work, to gain a better understanding of the effects of irregular motions, we try to search for optimal motion which can lead to high hydrodynamic efficiency of a flapping foil in uniform flow, and our candidate motion can be even non-periodic. 

In recent years, RL-based flow control has emerged as a rapidly growing research area in the fluid mechanics community \cite{BruntonStevenL2020MLfF,GarnierPaul2021Arod}. RL-based method enables the design of complex flow control strategies and also advances our understanding of animal behavior through its ability to mimic biological learning \cite{GazzolaMattia2014Rlaw,GazzolaM.2016Ltsi,ColabreseSimona2017FNbS,NovatiGuido2017Stlf,VermaSiddhartha2018Ecsb,TsangAlanChengHou2020Shts,JiaoYusheng2021Ltsi}. Although RL methods have been widely adopted in numerical simulations \cite{RabaultJean2019Annt,NovatiGuido2019Cgap,MirzakhanlooMehdi2020AciS,GhraiebH.2021Sdrl,DuraisamyKarthik2021Poml,BorraFrancesco2022Rlfp,QiuJingran2022Nomi,LiJichao2022Rcoc,ZhuGuangpu2022Olpv}, the experimental applications remain limited \cite{ReddyGautam2018Gsvr,FanDixia2020Rlfb}. In this research, we treat the current motion optimization process of a continuously flapping foil as a sequential decision-making problem and introduce a model-free reinforcement learning (RL) framework to solve it. In this framework, the foil can iteratively adjust its motion by interacting with water tunnel environment, and we let the agent explore in a much wider motion space. Our work demonstrates the effectiveness of RL method in designing bionic flow control strategy though training in experimental flow environment.

\section{EXPERIMENTAL SET-UP AND METHODOLOGY}
\subsection{Experimental set-up}
\begin{figure*}
\includegraphics[width=0.9\textwidth]{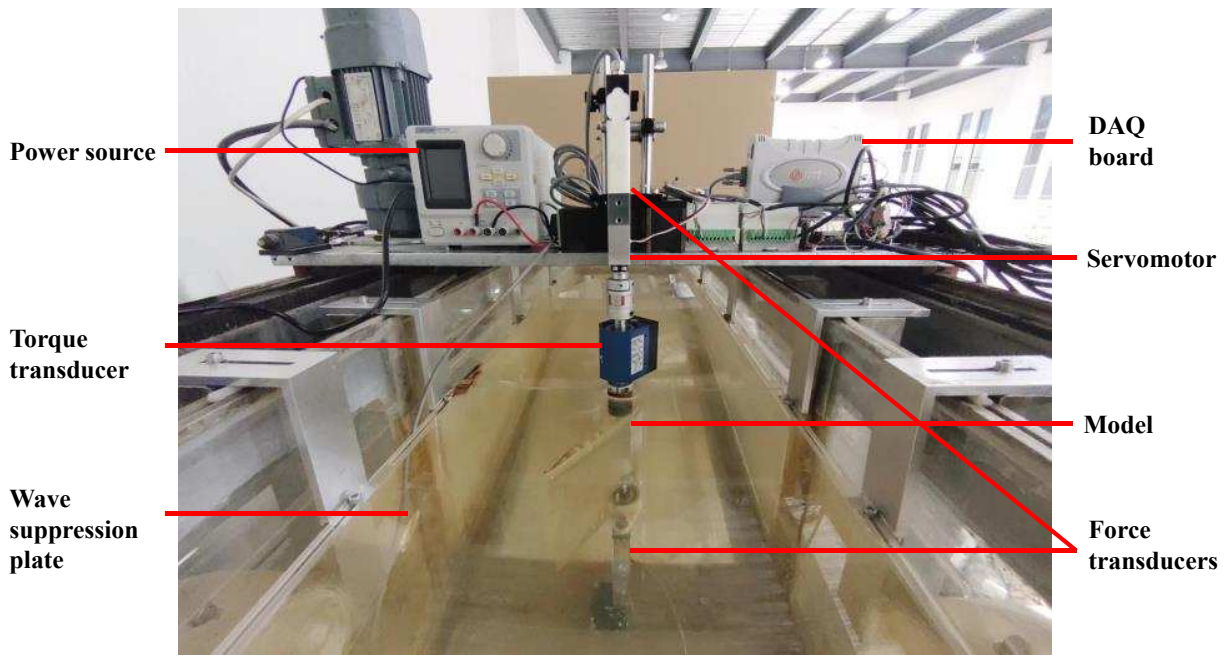}
\caption{\label{fig:1} Water tunnel and experimental instruments.}
\end{figure*}

Our experiment is carried out in a water tunnel with a test section of 0.5m (width) by 0.5m (height) by 6m (length), and the water speed range of the tunnel is 0-5m/s. The test model is a NACA0012 foil with chord length $c$ of 0.2m and span $s$ of 0.2m. The rotation axis of the model is set at 20\% chord length position from the leading edge. In this work, the water speed $U$ is set at 0.077m/s for all the tests, resulting in a chord-based Reynolds number of $Re_c = 13500$. To collect feedback information from the environment, two force transducers are mounted at both ends of the model, while the torque transducer is set at the top end, as shown in Fig. \ref{fig:1}. Besides, a serial bus servomotor (STS3046) is used to drive the foil model to realize arbitrary flapping motion. Load signals are collected on a data acquisition card (JY USB-62401) and synchronized with feedback motion signals collected from serial bus servo. Sampling rate for all the signals is set as 80Hz.

In this research, the propulsive performance of the flapping foil is evaluated by its thrust coefficient and Froude efficiency, conventionally defined as
\begin{equation}
C_T= \frac{T}{\frac{1}{2} \rho U^2 sc},  \quad 
\eta = \frac{W}{P} = \frac{TU}{M\omega}
\end{equation}
where $T$ is the net streamwise component of the hydrodynamic force induced by the flapping motion, while $\rho$ refers to the fluid density. $W$ and $P$ denote the power expended and developed by the motion respectively, while $M$ represents the torque exerted at the axis of rotation and $\omega$ is the angular velocity of rotation.

\subsection {Reinforcement Learning algorithm}
A reinforcement learning problem is typically modelled as a Markov Decision Process (MDP), denoted by $\mathcal{M} = <S,A,P,R,\gamma>$. In this setting, the agent takes an action $a_t\sim \pi_x (a|s_t)$ from a policy $\pi: S \times A \to [-1,1]$ parameterised by $x$, given an initial state $s_0 \in S$, and obtains a reward $r_{t+1} \sim R(s_t, a_t)$. The goal is to learn a policy that maximises the expected reward $\mathbb{E}[\sum_{t=0}^{\infty} \gamma^t r_t|s_0,\pi_x]$, under discount factor $\gamma \in [0,1)$.

In the context of current flapping foil system, the action of the agent is defined as the next tail-beat. A tail-beat refers to the process in which the trailing edge of the model moves from one end to the other. In this study, single tail-beat motion is prescribed as a half-period sinusoidal motion with given amplitude and frequency, and we force each tail-beat to cross the center line. Thus, the action variable of current problem is denoted by $a_t=\{A_t, f_t\}$.  To balance between thrust and efficiency, we limit the flapping amplitude on each side within the range of $7^\circ$ to $20^\circ$. Additionally, we set the frequency range of every single tail-beat motion such that the instantaneous Strouhal number ($St=2Af/U$) of each motion falls between 0.2 and 0.8. The agent can choose any combination of tail-beat amplitude and frequency within this action space to maximize its hydrodynamic efficiency.

To set up a Markov process under uniform incoming flow condition, we use motion history as the state variable rather than the information of the instantaneous flow field, in order to avoid the sensor delay in real-time measurement. Specifically, the state that the agent observes is the history of $n$ tail-beats, denoted by $s_t=\{a_{t-n+1} \cdots a_t\}$. In this study, the value of $n$ is chosen to ensure that the state window is long enough for water to flow past 5 times of foil chord length in the whole tail-beat motion space, so that the influence of earlier wake on the foil performance becomes negligible, thereby guaranteeing the Markov property of current problem.

\begin{figure}[hb]  \includegraphics[width=0.49\textwidth]{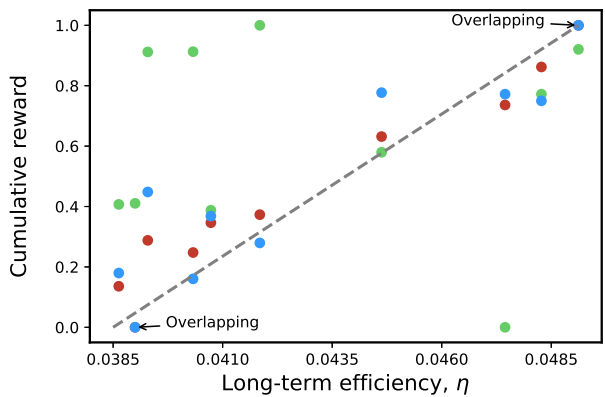}
  \caption{Mismatch between cumulative reward and long-term efficiency. Green, red, and blue dots represent the rewards calculated by $k=1$, $k=8$, and $k=16$ respectively, over ten one-minute episodes of random actions. The rewards under the same k value are normalized to the range of [0, 1].}
  \label{fig:mismatch}
\end{figure}

\subsection{Reward function}
Properly designing the reward function is critical in RL. When it comes to current efficiency optimization problems, we can simply set the efficiency of the action as the reward function in RL framework. However, using short-term efficiency as the reward can harm the final performance because of the mismatch between cumulative short-term efficiency and long-term efficiency. This issue can be illustrated as follows. Assuming $W_0, W_1, \cdots, W_{T-1}$ are useful works produced by $T$ continuous tail-beat motions of the foil, and $P_0, P_1, \cdots, P_{T-1}$ are the corresponding total works expended. If the reward is evaluated by hydrodynamic efficiency of $k$ continuous tail-beat motions, with assumption of $\gamma = 1$ for simplicity, the expected cumulative reward function of this whole T-length episode from $s_0$ can be expressed by $(W_0+\cdots+W_{k-1})/(P_0+\cdots+P_{k-1}) +(W_1+\cdots+W_k)/(P_1+\cdots+P_k) +\cdots+(W_{T-k}+\cdots+W_{T-1})/(P_{T-k}+\cdots+P_{T-1})$. This is the objective that RL agent aims to maximize. However, what we want to maximize in current problem is not just hydrodynamic efficiency of every single motion, but long-term efficiency, which is calculated by $(W_0+W_1+\cdots+W_{T-1})/(P_0+P_1+\cdots+P_{T-1})$. Obviously, to maximize a reward function defined by short-term efficiency might not lead to high long-term efficiency.

To further investigate this problem, ten one-minute episodes of random flapping motions are measured and the cumulative reward functions are calculated with different k values as shown in the last paragraph, and the result is presented in Fig. \ref{fig:mismatch}. When the efficiency of single action is used as the reward function (i.e., $k=1$), the linear relationship between cumulative reward and long-term efficiency is weak. As $k$ increases, the linearity becomes better. However, increasing the window length of reward evaluation also smooths out the reward space, which makes it challenging for the agent to identify actions that could potentially benefit the long-term efficiency and slows down the training process.  Therefore, it is necessary to consider both the optimization objective and the training feasibility when selecting the value of $k$. 

\begin{figure*}[htb]
  \centering \includegraphics[width=0.65\textwidth]{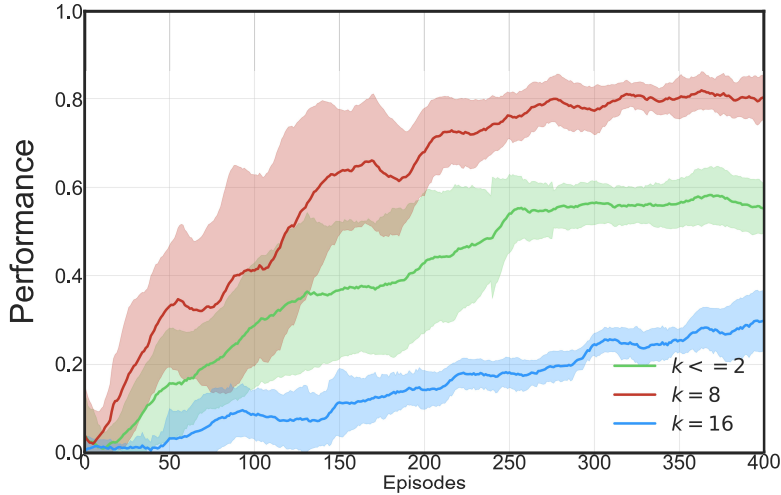} 
  \caption{Learning curves of different levels of $k$. The solid line and shaded area denote the mean and standard deviation respectively over five independent experiments. The performance has been normalized based on long-term efficiency, where performance of 1 and 0 corresponds to efficiency of 16$\%$ and 4$\%$ respectively. }
\label{fig:learning_curve}
\end{figure*}

\subsection {Experimental procedure}
At the beginning of each tail-beat, the agent makes a decision and takes action based on the current state. Prior to each episode, the foil performs its initial state motion twice to warm up the equipment and initialize the flow field. Following the initialization process, the agent interacts with the environment for one minute and all the data are collected. And to avoid interference from the flow field in successive experiments, a one-minute break is inserted after each episode. Notably, the entire experimental procedure is automated and iterated for hours in the water tunnel laboratory without manual monitoring before the training is accomplished.

\begin{figure*}
  \centering
  \begin{subfigure}{0.65\textwidth}
    \includegraphics[width=\textwidth]{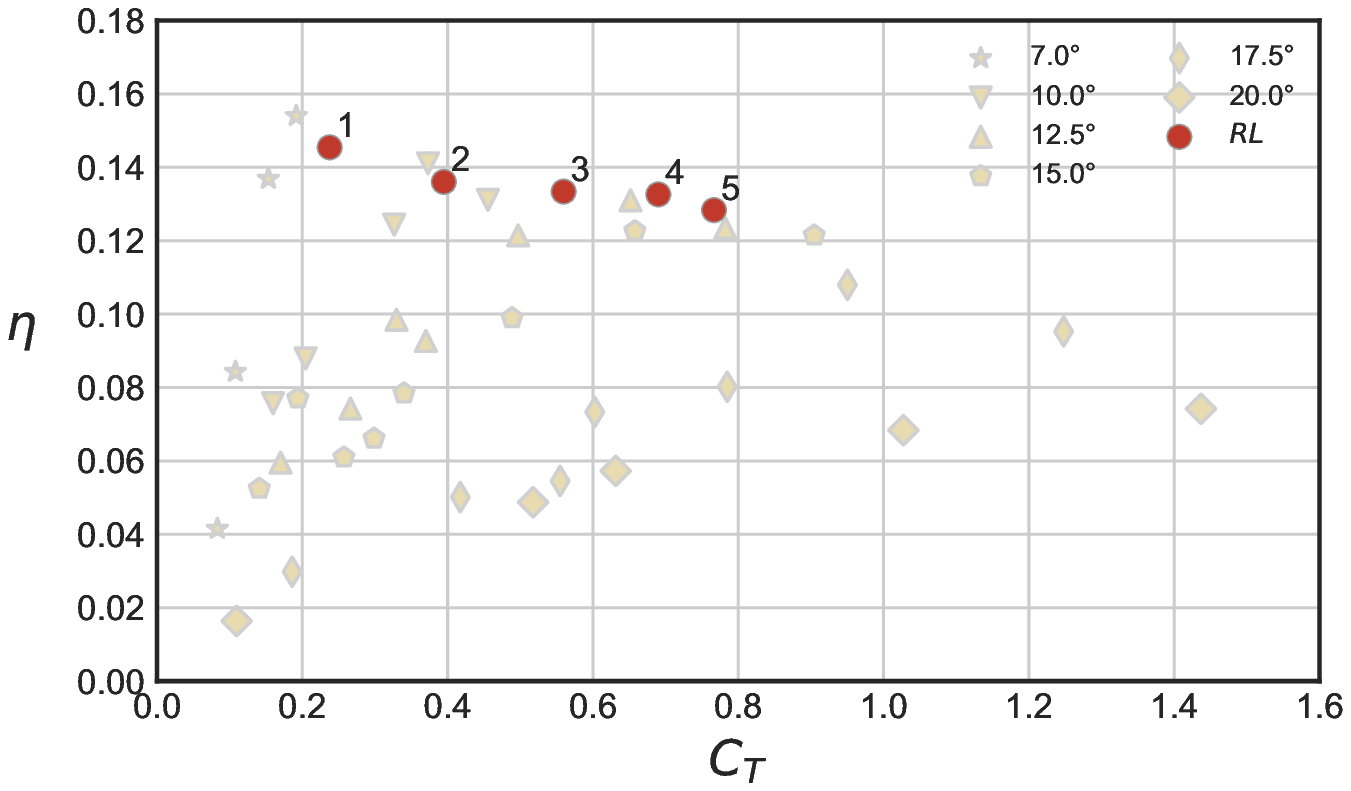}
    \caption{}
        \label{fig:sub2_a}
  \end{subfigure}
  \hfill
  \begin{subfigure}{0.49\textwidth}
    \includegraphics[width=\textwidth]{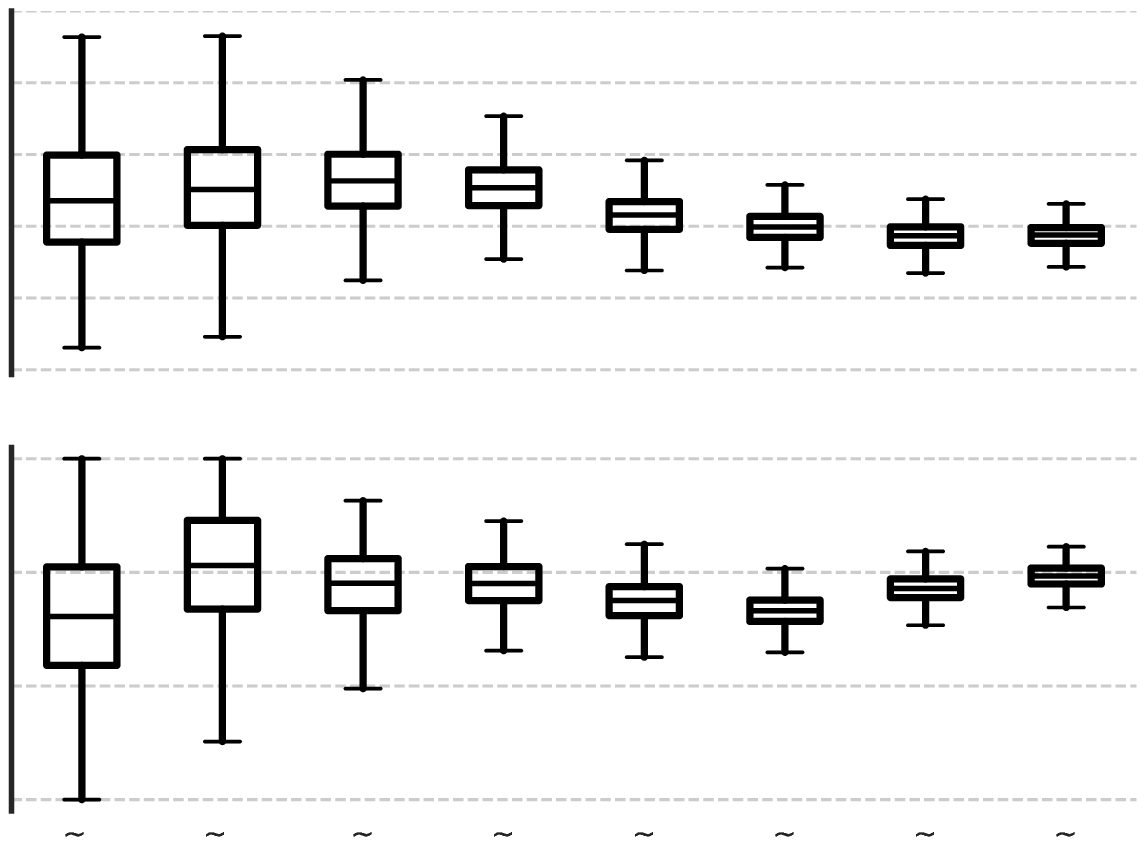}
    \caption{}
    \label{fig:sub2_b}
  \end{subfigure}
   \begin{subfigure}{0.49\textwidth}
\includegraphics[width=\textwidth]{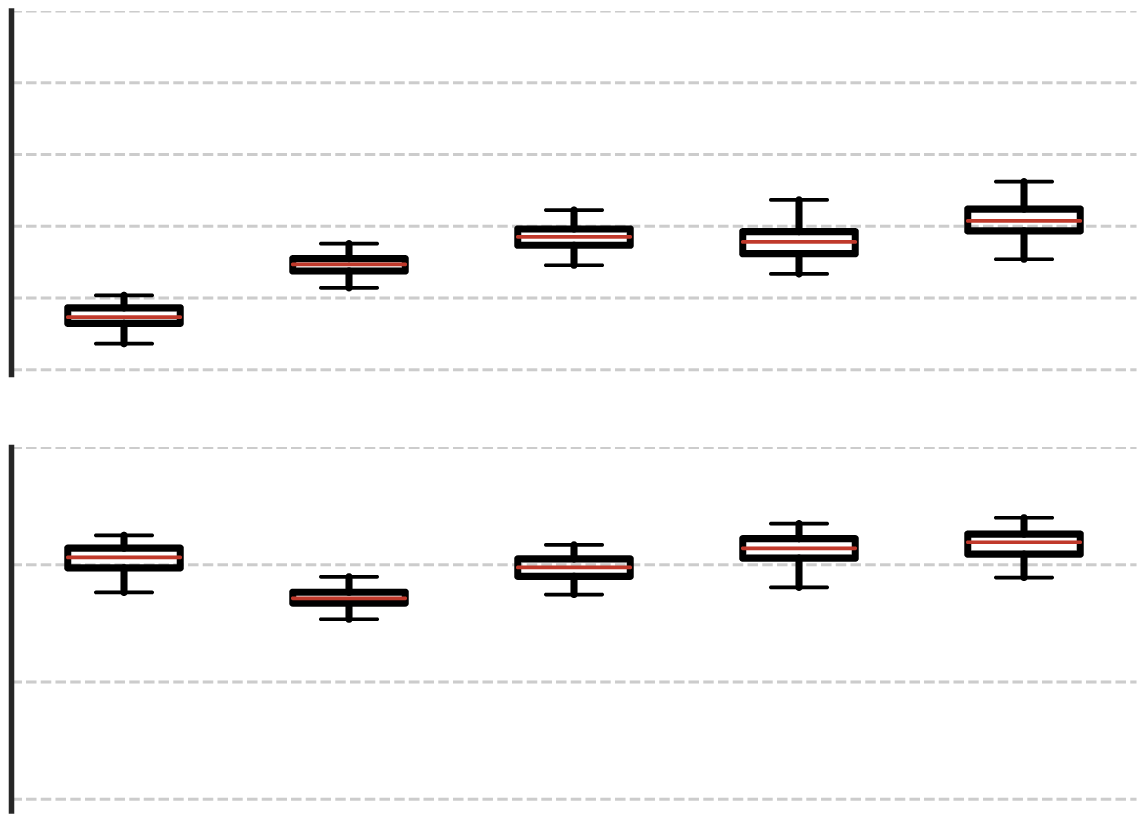}
    \caption{}
    \label{fig:sub2_c}
  \end{subfigure}
  \caption{Performance of the learned strategies. (a) The light yellow dots in the graph represent the sinusoidal motion family, while the red dots indicate the motion strategies of the RL agent. All sinusoidal motions are included in the action space of the current RL agent. (b) Learning path of Expt. 3. The median is shown as the midline in the box plot, while the extent of the boxes marks the interquartile range, and the whiskers demarcate the outliers (1.5 times of the interquartile range). (c) Learned swimming strategies of five independent experiment.
 }
  \label{fig:5}
\end{figure*}

\section{RESULTS}
In the results section, we begin by demonstrating that the number of tail-beats, denoted by $k$, in the efficiency evaluation has can directly influence the training process, as illustrated in Fig. \ref{fig:learning_curve}, and we discuss the underlying causes of this phenomenon. We then compare the motion pattern of the agent  to the sinusoidal motion family on the $C_T - \eta$ graph, as shown in Fig. \ref{fig:5}. All the sinusoidal motions drawn are contained in the action space of the current RL agent.

\subsection{The effect of reward function}
In the last section, we have discussed the discrepancy between long-term efficiency and the training objective of the RL agent, especially when the evaluation window of reward function is small. Here we investigate the influence of $k$ on the training process by the results of a series of experiments. Our study consists of five sets of experiments at $k\leq2$, $k=8$ and $k=16$ respectively. Specifically, we group three cases of $k=1$ and two cases of $k=2$ together due to their similar properties. It is also important to note that as the value of $k$ increases, the state window $n$ is also extended accordingly. 

As shown in Fig. \ref{fig:learning_curve}, when $k=8$, learning curve of the RL agent rises most rapidly and the final performance is also the best among the three groups. When reward evaluation window is small ($k\leq2$), the learning process is slower, since the short-term efficiency is highly sensitive to environmental noises, and the final performance is also worse than the results obtained when $k=8$ due to the discrepancy between short-term objective and long-term efficiency. On the other hand, when the reward evaluation window is too large ($k=16$), the RL agent seems to be trapped in a smooth reward zone and almost stagnates in the early stage of training process. And the learning curve rises very slow in the following episodes, since it is difficult for the agent to explore for more efficient strategy in a very long-term sense in the neighbourhood of its current motion. The agent might reach a good performance for $k=16$ with much more episodes of training, but it is too time consuming. Through this study, we finally determine the optimal value of $k$ for current problem is 8, as it offers the best long-term efficiency performance while requiring the least amount of training time.

\subsection{Comparison with sinusoidal gaits}
A comparison of the performance of flapping motions obtained from the current RL process with sinusoidal motions is particularly intriguing, given that no continuous flapping motion has been reported to be more efficient than sinusoidal motion. In this section, all the RL results under discussion are obtained with $k=8$, which is the best reward window setting for current problem. Hydrodynamic efficiency and thrust coefficient of sinusoidal motion with every group of given frequency and amplitude are calculated over 60-second period and averaged in 5 independent measurements. Hydrodynamic performance of RL motions are evaluated in the same way. Since the flapping motions obtained from RL agent are not periodic, we cannot describe these motions by Strouhal number. Thus, we compare RL motions and sinusoidal motions in a $C_T - \eta$ plot.

Interestingly, as shown in Fig. \ref{fig:sub2_a}, the RL agent consistently converges to the upper efficiency boundary of the sinusoidal motion family within 400 training episodes. To further analyze the learning path of each RL training process, we extract all the intermediate motion data from RL Expt. 3. We take motion time histories of every 50 training episodes and investigate the distribution of amplitude and frequency of all the tail-beats in these data sets, and the results are illustrated by boxplot in Fig. \ref{fig:sub2_b}. It can be observed that, at the onset of training, the boxplots are tall for both amplitude and frequency, and the agent simply makes its action decision randomly. As the training progresses, the agent begins to narrow the scope of the action space exploration, which is shown in Fig. \ref{fig:sub2_b} by the shrink of boxplots along the episode axis. In the latter period of training, the agent has learned to adjust its tail-beat motion within a narrow range of frequency and amplitude to acquire higher efficiency. Boxplots of the motion time histories of the final episodes from five independent RL experiments are plotted in Fig. \ref{fig:sub2_c}. All five boxplots in this figure are very short, which indicates that the optimal motions obtained from RL training are all close to sinusoidal motion with fixed amplitude and frequency, while the height of the boxplots shows the difference between the RL motion and pure sinusoidal motion. Moreover, the median frequency and amplitude of these RL motions lead to an Strouhal number range of 0.35 to 0.55, which aligns with the optimal Strouhal number of sinusoidal motions.

\section{CONCLUSION}
In this work, we have demonstrated the feasibility of using RL method to optimize the hydrodynamic performance of a flapping foil in a real-world fluid environment. The study of non-periodic flapping motions requires a significant amount of time and human-guided selection, making it nearly impossible to carry out by traditional methods. However, through RL, we successfully find swimming strategies that adjust frequency and amplitude of tail-beat motions in a narrow range, which exhibit similar long-term efficiency as sinusoidal motions.

Our study also reveals the deviation between the optimization goal and long-term efficiency when the evaluation window of reward function is small. When the reward function is not designed properly, the agent is likely to be trapped in a sub-optimal region. On the other hand, we also observe that when the reward function is calculated in a excessively large window, the training process remarkably slows down or even stagnate in certain period. However, this problem might be solved by additional techniques such as pre-training. 

In current research, RL method shows its capability in solving challenging problem brought by complex vortex flow field around a flapping foil. In future research, more motion patterns could be designed and investigated, such as introducing additional degrees of freedom or designing the swimmer to be more flexible like a real fish. With the assistance of RL training, the hydrodynamic performance of the swimmer is expected to be improved further. Additionally, environmental feedback variables could be added to the state vector of the current RL method, enabling the agent to adjust its decision based on real-time feedback from the environment and potentially providing the swimmer with adaptability in different flow environments.

\begin{acknowledgments}
We thank support from the department and the university.
\end{acknowledgments}

\section*{Data Availability Statement}

The data that support the findings of this study are available from the corresponding author upon reasonable request.

\appendix

\section{RL hyperparameters}
The policy network in this research consists of a long short-term memory (LSTM) layer with width of 64 and a hidden layer with width of 128, while the value network consists of a
LSTM layer and a hidden layer, both with width of 128. The discount factor $\gamma$ and learning rate is set as 0.999 and 0.0002 respectively. For hyperparameter settings in PPG algorithm, we use the settings recommended in \cite{CobbeKarl2020PPG}. 

All the RL training process in the text are started with the same initial parameters.

\nocite{*}

\bibliography{aipsamp}

\end{document}